\documentclass{amos}

\usepackage{amsmath}
\usepackage{amssymb}
\usepackage{bm}
\usepackage{xcolor}
\usepackage{hyperref}
\hypersetup{colorlinks=true, linkcolor=blue, citecolor=blue, urlcolor=blue, pageanchor=false}
\usepackage[numbers]{natbib}
\usepackage{siunitx}
\usepackage{booktabs}
\usepackage{tikz}
\usetikzlibrary{arrows.meta, positioning, shapes.geometric, calc, fit, backgrounds}
\usepackage{subcaption}
\usepackage[capitalize]{cleveref}
\crefname{figure}{Fig.}{Figs.}
\Crefname{figure}{Fig.}{Figs.}
\crefname{table}{Table}{Tables}
\crefname{section}{Section}{Sections}
\crefname{equation}{Equation}{Equations}

\AtBeginDocument{%
  \let\linkedciteauthor\citeauthor
  \renewcommand{\citeauthor}[1]{\begin{NoHyper}\linkedciteauthor{#1}\end{NoHyper}}%
}

\title{\TitleFont GPU-Accelerated Astrodynamics World Models \\ for Spacecraft Rendezvous and Proximity Operations}

\author{Duncan Eddy \\ Stanford University \and Isaac R. Ward, Grace Ra Kim, and Mykel J. Kochenderfer \\ Stanford University}

\date{}

\begin{document}

\maketitle

\begin{abstract}

World models are an emerging paradigm in representation learning in which an agent jointly learns state-action dynamics and observation models from offline trajectory data, enabling multi-step planning and trajectory prediction with uncertainty estimates. They have demonstrated strong results in robotics and game environments, but, to the best of our knowledge, they have not previously been applied to the space domain. This paper introduces a world model-based approach to cooperative and non-cooperative spacecraft rendezvous and proximity operations, and makes three contributions. First, we introduce an open-source, JAX-based International Space Station (ISS) docking simulation environment that supports parallel GPU-based simulation of spacecraft orbit and attitude dynamics, generating the thousands of state-action transitions that world model training requires. Second, we introduce \emph{Out-of-this-World-Model}, a transformer-based world model that encodes relative kinematic states and body-fixed camera imagery into a latent state and predicts its evolution under commanded thrusts and control torques using one-step flow matching. The model produces a distribution over future observations, capturing stochastic dynamics and providing per-timestep uncertainty estimates, and achieves greater predictive performance than DreamerV3-style posterior-correction baselines with fewer trainable parameters and hyperparameters. Third, we apply the approach to a capsule autonomously docking with the ISS under keep-out-zone constraints, demonstrating improved sample efficiency and task performance over reinforcement learning baselines ($53\%$ versus $29\%$ docking success across ports), substantially better out-of-distribution generalization (on held-out ports the world model more than doubles baseline success, $40\%$ versus $17\%$), and detection of anomalous objects encountered during the docking approach with $98\%$ classification accuracy. We open-source the simulation environment and model architecture to enable further study of this paradigm.

\end{abstract}

\section{Introduction}
\label{sec:introduction}

Spacecraft rendezvous and proximity operations (RPO) are transitioning from a rare, human-supervised procedure to a routine element of civil, military, and commercial space operations. Orbital rendezvous was first demonstrated in 1965 when Gemini~6A closed and held close proximity with Gemini~7. Rendezvous and docking remain essential to human space exploration today, with every crew and cargo vehicle visiting the International Space Station (ISS) performing one. These capabilities are currently central to future exploration architectures such as the Human Landing System, which requires multiple automated dockings and propellant transfers per mission~\cite{Chavers2020HLS}. The technology is also dual-use. There is also a growing tempo of military RPO activity in both low Earth orbit and the geostationary belt, with inspection and shadowing maneuvers by multiple nations becoming a recurring feature of the space domain~\cite{Swope2025space}. Beyond orbital games, on-orbit servicing is seen as a key enabler for reducing the cost of space operations by extending the life of satellites limited by consumables rather than hardware failures. The Orbital Express program demonstrated autonomous servicing in-orbit~\cite{Friend2008orbital}, and Northrop Grumman's Mission Extension Vehicles have since docked with operational geostationary communications satellites to provide life-extension services~\cite{Pyrak2022mev}.

Despite this growing operational tempo, rendezvous remains a complicated and risky procedure. Two spacecraft must be brought from kilometers of separation to physical contact with centimeter-level precision, and any collision risks permanent damage to both vehicles. A collision during approach is not only a mission failure---it is a potential debris-generating event that places other satellites in the orbital population at risk. Safe autonomous rendezvous therefore requires methods that can reason about the consequences of control actions before execution and recognize off-nominal scenarios early enough to initiate an abort before the operation leads to catastrophic failure.

For an autonomous rendezvous agent, this reasoning must span both uncertain dynamics and uncertain sensing. The agent observes the world through a range of potential sensors---GNSS receivers, cameras, star trackers, sun sensors, LIDAR, and ground-based tracking---and must fuse these measurements into an estimate of the relative state before deciding on a course of action. Operations can be cooperative, where the target vehicle exchanges state information and measurements with the chaser, or non-cooperative, where the chaser relies entirely on its own sensing. Cooperative measurements, typically GNSS observables, reduce relative state uncertainty significantly, but they can never remove the aleatoric noise of the underlying sensors. Cameras can provide measurements of relative position and attitude without relying on cooperative information exchange, making them robust to communications failures and supporting scenarios where cooperation is not possible, but provide less accurate relative state measurements. The central question for autonomy in either case is how to fuse the available, heterogeneous measurements and reason over their meaning when selecting actions.

Historically, this question has been answered by decomposing the problem into separate guidance, navigation, and control (GNC) functions. A filter---often a Kalman filter variant~\cite{grewal2010applications,crassidis2007survey}---combines an analytic dynamics model with analytic measurement models to produce a state estimate, then a guidance and control layer plans maneuvers against that estimate. Model predictive control has been applied to the docking problem with explicit constraint handling~\cite{Weiss2015mpc}, and artificial potential function methods provide safety guarantees under motion constraints~\cite{Dong2017safety}. These decompositions work well when the dynamics and measurement models are accurate, and the observations are low-dimensional. They offer no direct mechanism, however, for ingesting rich sensing modalities such as camera imagery, which must first be preprocessed through a computer vision pipeline to extract relative pose measurements before the filter can consume them.

Learning-based methods promise to close this gap by learning system behavior directly from data, but existing approaches fall short of the full problem. Directly regressing state dynamics or next observations with feedforward networks degrades rapidly over multi-step prediction horizons as errors compound \cite{lambert2022investigating}. Physics-informed neural networks encode dynamical constraints, such as conservation of energy or angular momentum, into the training loss, but they are brittle to train and model only the dynamics portion of the system evolution---they provide no mechanism for predicting the sensor observations that the agent actually receives. \citeauthor{Martin2023pinngm} study physics-informed gravity modeling and document these training challenges~\cite{Martin2023pinngm}. Reinforcement learning has been applied to spacecraft guidance more broadly~\cite{hovell2021deep,tipaldi2022learning}, but learns a reactive policy tied to the reward and scenario distribution it was trained on rather than a reusable model of the system.

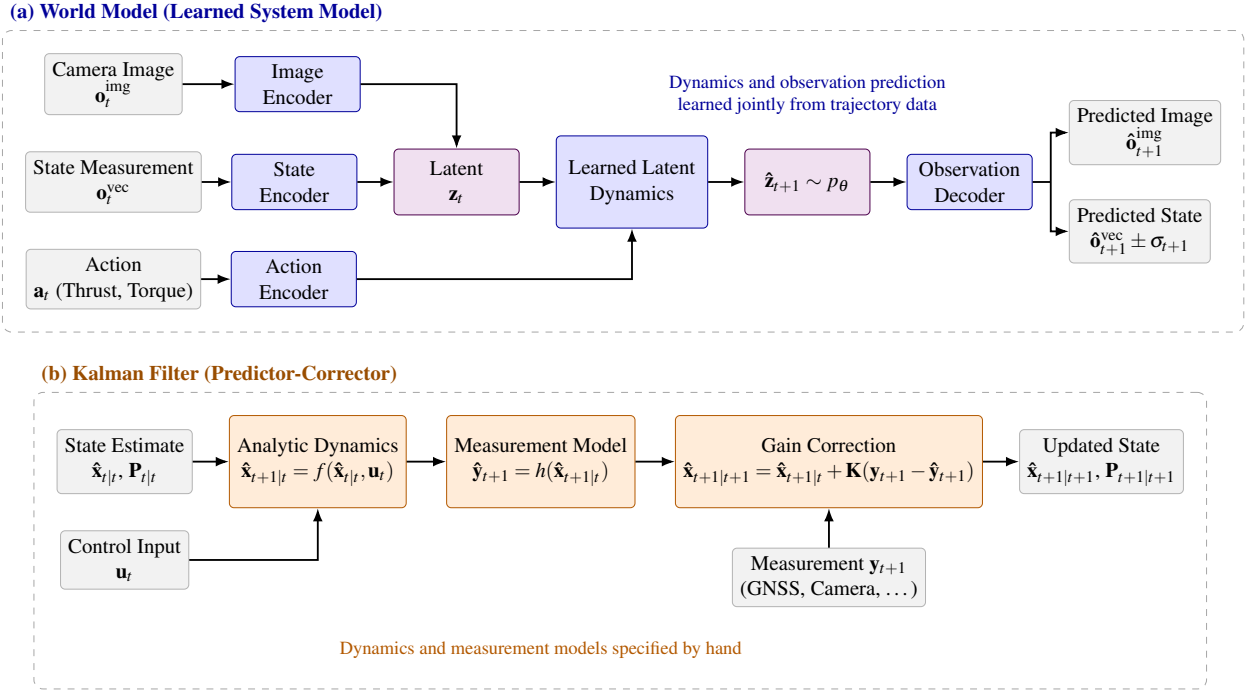
\begin{figure}[t]
    \centering
    \resizebox{\textwidth}{!}{
\begin{tabular}{@{}c@{}}
\begin{tikzpicture}[
    font=\small,
    node distance=0.55cm,
    box/.style={draw, rounded corners=2pt, minimum width=1.9cm, minimum height=0.75cm, align=center, fill=white},
    learned/.style={box, fill=blue!12, draw=blue!60!black},
    analytic/.style={box, fill=orange!15, draw=orange!70!black},
    data/.style={box, fill=gray!10, draw=gray!60, minimum width=1.8cm},
    latent/.style={box, fill=violet!12, draw=violet!60!black},
    arr/.style={-{Latex[length=2mm]}, thick},
    lbl/.style={font=\footnotesize, align=center},
    panel/.style={draw=gray!70, dashed, rounded corners=4pt, inner sep=10pt}
  ]

  \begin{scope}[local bounding box=wmpanel]
    \node[data] (img) {Camera Image\\$\mathbf{o}^{\text{img}}_t$};
    \node[data, below=of img] (statemeas) {State Measurement\\$\mathbf{o}^{\text{vec}}_t$};
    \node[data, below=of statemeas] (action) {Action\\$\mathbf{a}_t$ (Thrust, Torque)};

    \node[learned, right=0.79cm of img] (imgenc) {Image\\Encoder};
    \node[learned, right=0.45cm of statemeas] (vecenc) {State\\Encoder};
    \node[learned, right=0.45cm of action] (actenc) {Action\\Encoder};

    \node[latent, right=0.55cm of vecenc, minimum height=1.0cm] (zt) {Latent\\$\mathbf{z}_t$};
    \node[learned, right=0.55cm of zt, minimum width=2.3cm, minimum height=1.4cm] (dyn) {Learned Latent\\Dynamics};

    \node[latent, right=0.55cm of dyn, minimum height=1.0cm] (znext) {$\mathbf{\hat z}_{t+1} \sim p_\theta$};

    \node[learned, right=0.55cm of znext] (dec) {Observation\\Decoder};
    \node[data, right=0.55cm of dec, yshift=0.75cm] (predimg) {Predicted Image\\$\mathbf{\hat o}^{\text{img}}_{t+1}$};
    \node[data, right=0.55cm of dec, yshift=-0.75cm] (predvec) {Predicted State\\$\mathbf{\hat o}^{\text{vec}}_{t+1} \pm \mathbf{\sigma}_{t+1}$};

    \draw[arr] (img) -- (imgenc);
    \draw[arr] (statemeas) -- (vecenc);
    \draw[arr] (action) -- (actenc);
    \draw[arr] (imgenc.east) -| (zt.north);
    \draw[arr] (actenc.east) -| (dyn.south);
    \draw[arr] (vecenc) -- (zt);
    \draw[arr] (zt) -- (dyn);
    \draw[arr] (dyn) -- (znext);
    \draw[arr] (znext) -- (dec);
    \coordinate (decsplit) at ($(dec.east)!0.5!(predimg.west |- dec.east)$);
    \draw[thick] (dec.east) -- (decsplit);
    \draw[arr] (decsplit) |- (predimg.west);
    \draw[arr] (decsplit) |- (predvec.west);

    \node[lbl, above=0.4cm of znext, text=blue!60!black] {Dynamics and observation prediction \\learned jointly from trajectory data};
  \end{scope}
  \node[panel, fit=(wmpanel)] (wmbox) {};
  \node[anchor=south west, font=\small\bfseries, text=blue!60!black] at (wmbox.north west) {(a) World Model (Learned System Model)};

\end{tikzpicture}
 \\[0.75em]
\begin{tikzpicture}[
    font=\small,
    node distance=0.55cm,
    box/.style={draw, rounded corners=2pt, minimum width=1.9cm, minimum height=0.75cm, align=center, fill=white},
    learned/.style={box, fill=blue!12, draw=blue!60!black},
    analytic/.style={box, fill=orange!15, draw=orange!70!black},
    data/.style={box, fill=gray!10, draw=gray!60, minimum width=1.8cm},
    latent/.style={box, fill=violet!12, draw=violet!60!black},
    arr/.style={-{Latex[length=2mm]}, thick},
    lbl/.style={font=\footnotesize, align=center},
    panel/.style={draw=gray!70, dashed, rounded corners=4pt, inner sep=10pt}
  ]

  \begin{scope}[shift={(78pt,0pt)}, anchor=north west]
    \begin{scope}[local bounding box=kfpanel]
      \node[data] (est) at (0,-1.0) {State Estimate\\$\mathbf{\hat x}_{t|t}$, $\mathbf{P}_{t|t}$};
      \node[data, below=of est] (ctrl) {Control Input\\$\mathbf{u}_t$};

      \node[analytic, right=0.55cm of est, minimum height=1.4cm] (pred) {Analytic Dynamics\\$\mathbf{\hat x}_{t+1|t} = f(\mathbf{\hat x}_{t|t}, \mathbf{u}_t)$};

      \node[analytic, right=0.6cm of pred, minimum height=1.4cm] (measmod) {Measurement Model\\$\mathbf{\hat y}_{t+1} = h(\mathbf{\hat x}_{t+1|t})$};

      \node[analytic, right=0.6cm of measmod, minimum height=1.4cm] (corr) {Gain Correction\\$\mathbf{\hat x}_{t+1|t+1} = \mathbf{\hat x}_{t+1|t} + \mathbf{K}(\mathbf{y}_{t+1} - \mathbf{\hat y}_{t+1})$};

      \node[data, below=0.6cm of corr] (meas) {Measurement $\mathbf{y}_{t+1}$\\(GNSS, Camera, \ldots)};

      \node[data, right=0.55cm of corr] (out) {Updated State\\$\mathbf{\hat x}_{t+1|t+1}$, $\mathbf{P}_{t+1|t+1}$};

      \draw[arr] (est) -- (pred);
      \draw[arr] (ctrl.east) -| (pred.south);
      \draw[arr] (pred) -- (measmod);
      \draw[arr] (measmod) -- (corr);
      \draw[arr] (meas.north) -- (corr.south);
      \draw[arr] (corr) -- (out);

      \node[lbl, below=1.9cm of measmod, text=orange!70!black] {Dynamics and measurement models specified by hand};
    \end{scope}
    \node[panel, fit=(kfpanel)] (kfbox) {};
    \node[anchor=south west, font=\small\bfseries, text=orange!60!black] at (kfbox.north west) {(b) Kalman Filter (Predictor-Corrector)};
  \end{scope}

\end{tikzpicture}

\end{tabular}
}
    \caption{World models compared with the classical predictor-corrector approach to state estimation and prediction. (a) A world model learns encoders, latent dynamics, and an observation decoder jointly from trajectory data. It ingests raw measurements---camera imagery and noisy kinematic state---together with the applied action, and predicts a distribution over future observations, from which per-timestep uncertainty estimates follow directly. (b) A Kalman filter propagates a state estimate through analytic dynamics and measurement models, correcting the prediction with a gain-weighted measurement residual.}
    \label{fig:wm-vs-kf}
\end{figure}

World models are an emerging paradigm in representation learning that addresses this full problem. First introduced by \citeauthor{Ha2018world}~\cite{Ha2018world}, a world model co-learns the joint state-action dynamics and the observation model of a system from trajectory data, producing a learned simulator that predicts future observations conditioned on actions. For readers accustomed to spacecraft GNC, a world model can be understood as a learned system model---a fusion of the Kalman filter's dynamics and measurement models into a single network that ingests raw measurements and actions and predicts the distribution over future measurements directly. \Cref{fig:wm-vs-kf} illustrates this correspondence.

 World models have demonstrated strong results in robotics and game domains, enabling multi-step reasoning, model-based planning, and quantification of predictive uncertainty~\cite{Hafner2025dreamerv3, Ward2026foundational}. Unlike the Kalman filter, world models currently carry no optimality guarantees, but they inherit none of the filter's structural assumptions either: no linearization, no Gaussian noise model, and no requirement that observations be reduced to low-dimensional residuals before use.

This paper introduces a world model approach to spacecraft rendezvous and proximity operations. Our contributions are threefold:
\begin{enumerate}
    \item We introduce \textit{AstroJAX}, an open-source astrodynamics framework written in JAX that supports massively parallel, GPU-accelerated simulation of orbit and attitude dynamics, along with an open-source ISS-docking simulation environment built on it for generating world model training data at scale.
    \item We introduce \textit{Out-of-this-World-Model} (OWM), a transformer-based world model architecture that fuses kinematic and visual observations into a shared latent space and predicts distributions over future observations with a one-step flow-matching head. To the best of our knowledge, this is the first application of world models to space operations.
    \item We demonstrate that the learned world model composes with classical sampling-based planning, specifically model predictive path integral control~\cite{Williams2017mppi}, to dock with the ISS, raising docking success across all eight ports from $29\%$ with reinforcement learning to $53\%$, an $84\%$ relative gain, with the margin widest at berthing ports never seen in the training data, where it more than doubles the success rate of goal-conditioned baselines ($40\%$ versus $17\%$) and reaches ports the baselines fail to acquire at all. At the same time the model's predictive uncertainty detects anomalies at runtime, correctly classifying approach sequences with docked vehicles not seen during training from approach sequences without such vehicles $98\%$ of the time.
\end{enumerate}
We open-source the simulation environments, model architecture, and training datasets to enable further study of this paradigm.\footnote{The model architecture and simulation environments are available at \url{https://github.com/sisl/outofthisworldmodel} and \url{https://github.com/sisl/outofthisworldmodel-envs}. Training datasets are published at \url{https://huggingface.co/sislaboratory}.}

The paper proceeds as follows. \Cref{sec:background} provides background on world models and reviews prior work on learning-based rendezvous and proximity operations. \Cref{sec:methods} presents the world model formulation, the OWM architecture and training procedure, and the GPU-accelerated astrodynamics simulation used to generate training data. \Cref{sec:experiments} describes the experimental setup, baseline algorithms, and evaluation criteria. \Cref{sec:results} presents results on rendezvous performance, generalization to unseen docking ports, and anomaly detection. \Cref{sec:conclusions} concludes, summarizing the work and identifying areas for future development.

\section{Background}
\label{sec:background}

This section reviews the two bodies of work that this paper brings together: world models from the representation learning literature, and methods for rendezvous and proximity operations from the spacecraft GNC literature.

\subsection{World Models}
\label{sec:background-world-models}

\citeauthor{Ha2018world} introduce the modern world model formulation as a generative recurrent network that learns a compressed spatial and temporal representation of an environment from recorded trajectories. They show that policies trained entirely inside the learned model transfer back to the real environment~\cite{Ha2018world}. The Dreamer line of work extends this formulation into a recurrent state-space model that learns a posterior over latent states from observations and a prior that predicts the latent forward in time, correcting the prior toward the posterior during training. The third generation of this architecture masters a wide range of game environments from Atari to Minecraft with a single set of hyperparameters~\cite{Hafner2025dreamerv3}. These posterior-correction architectures are the closest learned analog to the predictor-corrector structure of a Kalman filter, and we use a Dreamer-style model as an architectural point of comparison in this work.

A second family of world models discards observation reconstruction entirely. \citeauthor{LeCun2022path} proposes the joint-embedding predictive architecture (JEPA), which predicts future representations in latent space rather than future observations in data space~\cite{LeCun2022path}, and \citeauthor{Assran2023ijepa} demonstrate the approach at scale for images~\cite{Assran2023ijepa}. Latent-space prediction is cheaper and avoids modeling pixel-level detail irrelevant to control, but it admits a degenerate solution in which the encoder collapses to a constant. The literature offers several remedies. \citeauthor{Grill2020byol} stabilize latent prediction with a slowly updated exponential-moving-average target encoder~\cite{Grill2020byol}, and \citeauthor{Bardes2022vicreg} regularize the batch latent distribution directly with variance and covariance penalties~\cite{Bardes2022vicreg}. \citeauthor{Maes2026leworldmodel} show that a sketched isotropic-Gaussian regularizer yields stable end-to-end joint-embedding world models from pixels~\cite{Maes2026leworldmodel}. The architecture we introduce in \Cref{sec:methods} supports these collapse-prevention strategies as interchangeable components of a shared transformer backbone.

The final ingredient of our approach comes from generative modeling. Flow matching learns a velocity field that transports samples from a noise distribution to a data distribution along near-straight paths~\cite{Lipman2023flow, Liu2023rectified}, and shortcut models add a self-consistency objective that lets the same network take one large integration step instead of many small ones~\cite{Frans2025shortcut}. Applied to latent prediction, these methods produce a distribution over next latent states that can be sampled in a single network evaluation---fast enough to sit inside a sampling-based planner---while retaining the ability to represent multimodal and stochastic dynamics that a point-prediction architecture cannot.

Learned system models have begun to see application in domains adjacent to spacecraft operations. An early line of work applies the paradigm to high-dimensional fluid dynamical simulation: \citeauthor{Morton2018nips} learn the forced and unforced dynamics of unsteady fluid flows directly from computational fluid dynamics data and suppress vortex shedding by performing model predictive control over the learned model~\cite{Morton2018nips}, with subsequent work inferring distributions over the learned dynamics for uncertainty-aware control~\cite{Morton2019koopman} and pairing learned compressions with learned dynamics to recover missing data from high-order flow simulations~\cite{Carlberg2019recovering}. Closer to our setting, \citeauthor{Ward2026foundational} train a probabilistic world model in the latent space of a pretrained video tokenizer and show that its predictive uncertainty reliably detects failures in real-world robotic bimanual manipulation tasks~\cite{Ward2026foundational}---evidence that world model uncertainty is a practical runtime monitoring signal, a property we exploit for anomaly detection in orbit. To the best of our knowledge, however, no prior work applies world models to problems in the space domain.

\subsection{Rendezvous and Proximity Operations}
\label{sec:background-rpo}

Reinforcement learning is the most studied learning-based approach to RPO. \citeauthor{Broida2019rendezvous} apply proximal policy optimization to rendezvous guidance in cluttered orbital environments~\cite{Broida2019rendezvous}, and \citeauthor{Gaudet2020angleonly} develop reinforcement meta-learning for angles-only intercept and adaptive guidance problems~\cite{Gaudet2020angleonly, Gaudet2020adaptive}. \citeauthor{Federici2021deep} compare deep learning techniques for autonomous proximity operations guidance under image-based navigation~\cite{Federici2021deep}, and \citeauthor{Fereoli2025meta} extend meta-reinforcement learning to proximity operations in cislunar space~\cite{Fereoli2025meta}. \citeauthor{Allen2023spacegym} release SpaceGym, a suite of non-cooperative space game environments intended to spur reinforcement learning research in the domain~\cite{Allen2023spacegym}.

Much of this work is enabled by dedicated simulation infrastructure. \citeauthor{Stephenson2024bskrl} build the BSK-RL library of reinforcement learning environments on top of the Basilisk astrodynamics library~\cite{Stephenson2024bskrl}. \citeauthor{Stephenson2025inspection} use these tools to learn autonomous inspection policies with optimization-based safety guarantees~\cite{Stephenson2025inspection}, and \citeauthor{Herrmann2024smallbody} apply reinforcement learning to small-body science operations~\cite{Herrmann2024smallbody}. \citeauthor{HutererPrats2025rl} apply reinforcement learning to space-to-space surveillance scheduling~\cite{HutererPrats2025rl}. These simulation frameworks execute on CPUs. Our work is complementary: by expressing the full dynamics, sensing, and reward pipeline in JAX, we can generate the hundreds of thousands of transitions that world model training requires directly on GPU hardware in parallel.

Docking has also been studied extensively through the lens of classical control. \citeauthor{Weiss2015mpc} apply model predictive control to rendezvous and docking with explicit handling of thrust limits, approach-cone, and soft-docking constraints~\cite{Weiss2015mpc}. \citeauthor{Dong2017safety} construct artificial potential functions that guarantee safety under motion constraints~\cite{Dong2017safety}, and \citeauthor{Breeden2022guaranteed} provide docking safety guarantees with control barrier functions~\cite{Breeden2022guaranteed}. \citeauthor{Maestrini2022guidance} develop guidance strategies for inspecting unknown non-cooperative objects~\cite{Maestrini2022guidance}. These methods provide strong guarantees against the models they are given, but they inherit the limits of those models: rich observations such as camera imagery enter only after being processed by a computer vision pipeline to derive relative-state estimates.

A smaller body of work learns spacecraft dynamics directly. \citeauthor{Silvestrini2022deep} survey neural-network approaches to spacecraft dynamics, navigation, and control, and document the difficulty of maintaining accuracy over long propagation horizons~\cite{Silvestrini2022deep}. \citeauthor{Martin2023pinngm} develop physics-informed neural network gravity models and detail the training challenges of the approach~\cite{Martin2023pinngm}. These efforts model the dynamics alone. A world model differs in scope: it jointly learns the dynamics and the observation process, which is what allows it to consume camera imagery and kinematic measurements directly and to predict both forward in time.

\section{Methods}
\label{sec:methods}

Our approach has two components: a world model that learns the joint evolution of spacecraft state and sensor observations from trajectory data, and a GPU-accelerated simulation stack that generates that data at the scale world model training requires. This section presents the problem formulation, the Out-of-this-World-Model architecture and its training procedure, and the astrodynamics simulation environments.

\subsection{Problem Formulation}
\label{sec:methods-formulation}

We model the docking scenario as a discrete-time partially observable decision process~\cite{Kochenderfer2022algorithms}. At each timestep $t$, the environment occupies a state $\mathbf{s}_t$, the agent applies an action $\mathbf{a}_t$, and the state evolves according to a transition function
\begin{equation}
    \mathbf{s}_{t+1} = T(\mathbf{s}_t, \mathbf{a}_t)
    \label{eq:transition}
\end{equation}
where $T$ integrates the vehicle dynamics over one control interval. The agent does not observe $\mathbf{s}_t$ directly. Instead it receives an observation
\begin{equation}
    \mathbf{o}_t = \left(\mathbf{o}^{\text{vec}}_t, \mathbf{o}^{\text{img}}_t\right)
    \label{eq:observation}
\end{equation}
composed of a kinematic measurement $\mathbf{o}^{\text{vec}}_t$, the relative state corrupted by sensor noise, and a body-fixed camera image $\mathbf{o}^{\text{img}}_t$. The environment also emits a scalar reward $r_t$ used by the reinforcement learning baselines; the world model itself never consumes it.

The world model's task is to approximate the distribution over the next observation conditioned on a history of past observations and actions. Given a history window of length $H$ discrete time steps, the model input is a sequence of past observations and actions
\begin{equation}
    \mathbf{h}_t = \left(\mathbf{o}_{t-H+1}, \mathbf{a}_{t-H+1}, \ldots, \mathbf{o}_t, \mathbf{a}_t\right)
    \label{eq:history}
\end{equation}

The model then learns
\begin{equation}
    p_\theta\!\left(\mathbf{o}_{t+1} \mid \mathbf{h}_t\right)
    \label{eq:wm-objective}
\end{equation}
from a dataset of recorded transitions. This is the learned analog of the paired dynamics and measurement models of a navigation filter: a single network that ingests raw measurements and actions (control inputs) and predicts the distribution over what the sensors will report next. Because the prediction is a distribution rather than a point, the model represents both the stochasticity of the dynamics and the noise of the sensors, and the spread of its samples provides a per-timestep uncertainty estimate.

\subsection{World Model Architecture}
\label{sec:methods-architecture}

\begin{figure}[t]
    \centering
    \resizebox{\textwidth}{!}{\includegraphics{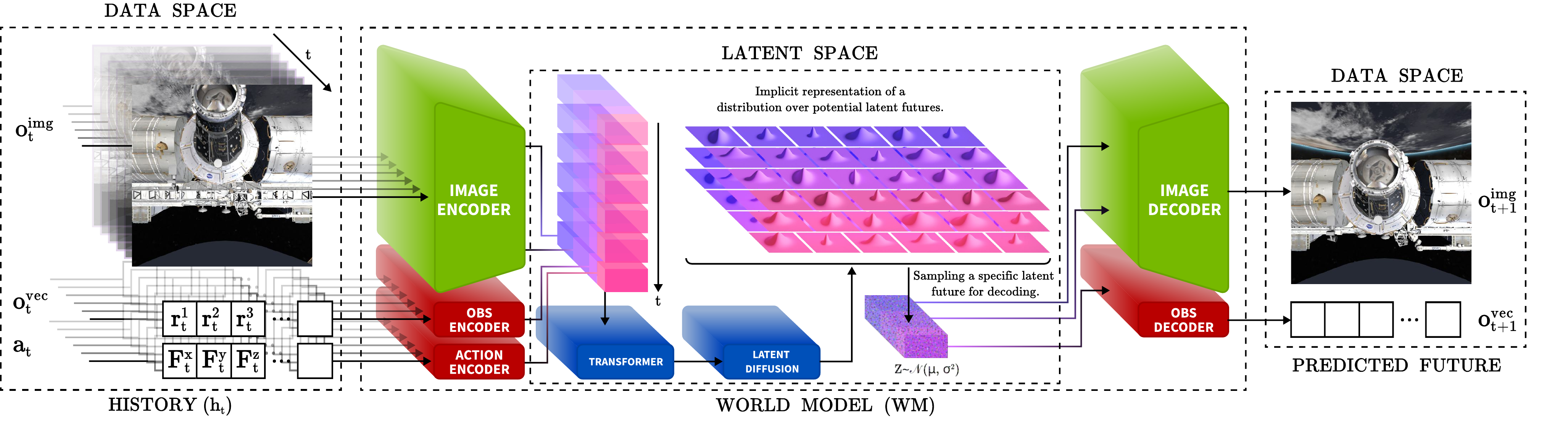}}
    \caption{The Out-of-this-World-Model architecture. A history window $\mathbf{h}_t$ of camera frames, kinematic state measurements, and actions is encoded into a vector of latent tokens per timestep. A factorized space--time transformer attends across tokens within each timestep and causally across timesteps. A flow-matching head transports Gaussian noise to a sample of the latent residual $\Delta\mathbf{\hat z}_{t+1}$; added to the current latent this gives the next latent state $\mathbf{\hat z}_{t+1} = \mathbf{z}_t + \Delta\mathbf{\hat z}_{t+1}$, and repeated draws with different noise realizations produce a distribution over futures. The sampled latent is appended to the token history and the model rolls forward autoregressively in latent space; an observation decoder maps any latent back to a predicted image and state with an associated uncertainty.}
    \label{fig:architecture}
\end{figure}

\Cref{fig:architecture} presents the Out-of-this-World-Model architecture. The design is modality-parameterized: each input stream contributes a fixed number of tokens to a per-timestep token vector, and the same backbone serves any combination of streams. A lightweight multilayer perceptron projects the kinematic state measurements, and a vision transformer encodes each camera frame into image tokens through a learned attention bottleneck that cross-attends a small set of latent queries against the frame's patch embeddings. The concatenation of the state and image tokens for each timestep is the model's latent state $\mathbf{z}_t$. A second multilayer perceptron projects the action into an additional token that is appended to the same per-timestep vector as conditioning; it is fused with the state and image tokens by the backbone's spatial attention.

The backbone is a factorized space--time transformer in the style of video architectures. Each block applies spatial attention---bidirectional attention among the tokens of a single timestep, which lets state, action, and image information fuse---followed by temporal attention, which attends causally across timesteps within a sliding window using rotary position embeddings. Factoring the attention this way reduces the cost from quadratic in the full token sequence to quadratic in each axis separately. A key--value cache makes autoregressive rollouts linear in horizon length, which addresses throughput bottlenecks when the model is queried inside a sampling-based planner.

Prediction is performed in latent space with a flow-matching head~\cite{Lipman2023flow, Liu2023rectified}, applied independently to each token of the vector. Conditioned on the backbone output for the current timestep, the head learns a velocity field that transports a standard Gaussian sample to the residual between the next latent state and the current one. The sampled latent is appended to the token history and the model rolls forward---latent-space autoregression---without ever decoding to observations in the loop. A decoder head per modality maps latents back to predicted images and states when observation-space output is needed.


\subsection{Training}
\label{sec:methods-training}

The model trains on windows of $H$ context steps followed by $F$ rollout steps sampled from the recorded trajectories. The flow head is trained with the rectified flow objective on the latent residual $\Delta\mathbf{z}_{t+1} = \mathbf{z}_{t+1} - \mathbf{z}_t$. For a noise sample $\mathbf{\epsilon} \sim \mathcal{N}(0, I)$ and a noise level $\tau \sim \mathcal{U}(0, 1)$, where $0$ is no added noise and $1$ is pure Gaussian noise, the noised input
\begin{equation}
    \mathbf{x}_\tau = (1 - \tau)\,\Delta\mathbf{z}_{t+1} + \tau\,\mathbf{\epsilon}
    \label{eq:flow-interp}
\end{equation}
lies on the straight path between data and noise, and the head $f_\theta$ minimizes
\begin{equation}
    \mathcal{L}_{\text{flow}} = \left\lVert f_\theta\!\left(\mathbf{x}_\tau, \tau, \mathbf{h}_t\right) - \left(\mathbf{\epsilon} - \Delta\mathbf{z}_{t+1}\right)\right\rVert^2,
    \label{eq:flow-loss}
\end{equation}
which regresses the constant velocity of that path.

Two families of reconstruction terms ground the latent space in observations. Let $g_m$ denote the decoder head for modality $m$, $\hat{\mathbf{z}}_t$ the latent the model predicts at step $t$, and $\mathbf{z}_t = E(\mathbf{o}_t)$ the encoding of the true observation $\mathbf{o}_t$, whose modality-$m$ component is $\mathbf{o}^{m}_t$. The per-modality decode loss reconstructs each observation stream from the \textit{predicted} latent,
\begin{equation}
    \mathcal{L}^{\text{dec}}_{m} = \frac{1}{|\mathcal{S}|}\sum_{t \in \mathcal{S}} \bigl\lVert g_m(\hat{\mathbf{z}}_t) - \mathbf{o}^{m}_t \bigr\rVert^2,
    \label{eq:loss-dec}
\end{equation}
where $\mathcal{S}$ is a random subset of the rollout steps drawn each iteration to bound decoder compute (the camera image is decoded by a deterministic mean-squared-error decoder; the kinematic state by its decode head). The latent round-trip anchor instead reconstructs the \textit{true} observation from its own encoding, with no dynamics in the loop,
\begin{equation}
    \mathcal{L}^{\text{rt}}_{m} = \frac{1}{T}\sum_{t} \bigl\lVert g_m(\mathbf{z}_t) - \mathbf{o}^{m}_t \bigr\rVert^2,
    \label{eq:loss-rt}
\end{equation}
which keeps the token codec close to an identity map so the shared latent stays faithful to the raw observations; we apply it to the image stream.

All terms combine as a weighted sum,
\begin{equation}
    \mathcal{L} = \lambda_{\text{flow}}\,\mathcal{L}_{\text{flow}}
                + \sum_{m} w_m\,\mathcal{L}^{\text{dec}}_{m}
                + \sum_{m} \beta_m\,\mathcal{L}^{\text{rt}}_{m},
    \label{eq:loss-total}
\end{equation}
with $\lambda_{\text{flow}} = 1$ and unit decode weights $w_m = 1$; the image round-trip anchor is weighted $\beta = 10$ (streams without an anchor take $\beta_m = 0$), since at unit weight it was a negligible fraction of the objective.

Training rolls the model forward autoregressively on its own predictions, with a teacher-forcing probability annealed from one to zero over the first epochs of training so the model learns to consume its own outputs. We train with AdamW at a learning rate of \num{e-3} with weight decay of \num{e-4}, a linear warmup over the first 300 steps, mixed \texttt{bfloat16} precision, and gradient-norm clipping at 1.0. Validation runs in fully autoregressive mode with deterministic sampling, so the monitored metrics reflect deployment conditions rather than teacher-forced ones. The models trained in this work use a token dimension of 128, a four-block space--time transformer backbone with eight attention heads and a 32-step sliding attention window, together with a separate two-block, four-head rectified-flow latent denoiser, totaling approximately 6.4M trainable parameters. On a single NVIDIA H100 an epoch takes 1.5 hours, with model convergence usually happening within the first 16 epochs.

\subsection{GPU-Accelerated Astrodynamics Simulation}
\label{sec:methods-astrojax}

World models typically require hundreds-of-thousands to millions of state-action transitions for successful training, which places a premium on simulation throughput. We generate training data with \texttt{outofthisworldmodel-envs}\footnote{Available at \url{https://github.com/sisl/outofthisworldmodel-envs}}, an open-source ISS-docking simulation environment whose dynamics, sensing, reward, and episode logic are written entirely in JAX~\cite{Bradbury2018jax}, so that environments can execute in parallel on CPUs or GPUs and entire trajectory rollouts compile to single fused programs. GPU parallelization enables significantly more throughput than is easily achievable with CPU-only simulation. The environment exposes a standard Gymnasium interface~\cite{Towers2024gymnasium} for direct integration with the wider reinforcement learning ecosystem, including a natively vectorized variant whose step, reset, observation, and reward paths are all batched rather than dispatched to parallel copies of a single environment.

\subsubsection{AstroJAX}
\label{sec:methods-astrojax-lib}

Many learning-based approaches rely on simple Hill--Clohessy--Wiltshire (HCW) dynamics to reduce computation and maximize simulation throughput, however this removes the more complex perturbations that shape relative motion over longer periods. Models that capture these perturbations may generate better policies, and, as a result, an astrodynamics simulation framework capable of high-throughput, high-fidelity parallel simulation is a key enabling technology for policy improvement and learning. To address this need, we developed \textit{AstroJAX}\footnote{Available at: \url{https://github.com/duncaneddy/astrojax}}, an open-source astrodynamics library that expresses standard astrodynamics models as pure, differentiable JAX functions. \Cref{tab:astrojax-models} inventories the implemented models.

\begin{table}[t]
    \centering
    \caption{Models implemented in AstroJAX.}
    \label{tab:astrojax-models}
    \begin{tabular}{@{}ll@{}}
        \toprule
        Category & Models \\
        \midrule
            Gravity & Point-mass, spherical harmonic, polyhedral \\
        Perturbations & Atmospheric drag (Harris--Priester, NRLMSISE-00), third-body Sun and Moon, \\
        & SRP with cylindrical or conical shadow \\
        Ephemerides & Analytic Sun and Moon, JPL approximate planetary positions \\
        Attitude & Quaternion kinematics, rigid-body Euler dynamics, gravity-gradient torque \\
        Relative motion & Hill--Clohessy--Wiltshire dynamics, \\
        & ECI--RTN transforms, quasi-nonsingular relative orbital elements \\
        Frames \& time & IAU 2006/2000A GCRF--ITRF with full Earth orientation data support, \\
        & TEME, geocentric, and geodetic coordinates, \\
        Propagation & RK4, RKF4(5), Dormand--Prince 5(4), RKN12(10) integrators; SGP4/SDP4 \\
        \bottomrule
    \end{tabular}
\end{table}

We validate AstroJAX against the Rust-backed brahe astrodynamics library~\cite{Eddy2026brahe}. The frame transformations reproduce IAU SOFA reference values to $10^{-8}$, and the gravity, third-body, solar radiation pressure, and drag accelerations agree with brahe to relative tolerances between $10^{-9}$ and $5\times10^{-4}$ depending on the model. The one exception is NRLMSISE-00 atmospheric density model implementation, needed for drag perturbation modeling, where the comparison is currently bounded near 15\% agreement.

\subsubsection{ISS Docking Environment}
\label{sec:methods-env}

The environment suite models a Dragon-class chaser maneuvering in the vicinity of the ISS. Three environments share a single task layer---the reward, docking criteria, collision geometry, sensor models, and reset distributions are identical---and differ only in the fidelity of their equations of motion. The results in this paper use the highest-fidelity member, \texttt{iss-numerical}, a full numerical simulation of the two-vehicle system.

In \texttt{iss-numerical}, the chief and chaser are propagated as independent state vectors in the Earth-centered inertial (ECI) frame through the same perturbed force model, which comprises zonal gravity harmonics through degree four (terms through $J_6$ are available), third-body accelerations from the Sun and Moon, and atmospheric drag with Harris--Priester density. The chaser's attitude is propagated inertially with quaternion kinematics and rigid-body Euler dynamics. The full simulation state is 21-dimensional---the epoch (defined by Modified Julian Day epoch and elapsed seconds), the chief and chaser ECI positions and velocities, the body-to-inertial quaternion, and the body rates---and integrates with a fourth-order Runge--Kutta scheme at a timestep of \SI{0.05}{\second} in double precision, with episodes running at most 7200 steps (\SI{360}{\second}). The chief flies an ISS reference orbit with a \SI{6795}{\kilo\meter} semi-major axis at \SI{51.64}{\degree} inclination. The task layer operates on a canonical 13-dimensional relative view derived from this state,
\begin{equation}
    \mathbf{s} = \left(\mathbf{r}, \mathbf{v}, \mathbf{q}, \mathbf{\omega}\right)
    \label{eq:state}
\end{equation}
comprising the relative position $\mathbf{r}$ and velocity $\mathbf{v}$ of the chaser with respect to the station origin in the rotating station-centered frame, the body-to-world attitude quaternion $\mathbf{q}$, and the body-frame angular velocity $\mathbf{\omega}$. The control
\begin{equation}
    \mathbf{a} = \left(\mathbf{F}, \mathbf{\tau}\right)
    \label{eq:control}
\end{equation}
is six-dimensional, comprising a body-frame force $\mathbf{F}$ and torque $\mathbf{\tau}$. Actuator limits are sized to Draco-class reality: a proximity-operations translation fires roughly four of Dragon's sixteen \SI{400}{\newton} thrusters along a body axis, giving a per-axis force limit of \SI{1600}{\newton}, and a thruster couple at a \SI{2.5}{\meter} arm gives a torque limit of \SI{2000}{\newton\meter}. The chaser has a mass of \SI{12000}{\kilogram} and principal inertias of \SIlist{80000;80000;50000}{\kilogram\meter\squared}.

Collision checking sweeps the chaser's \SI{2.25}{\meter} bounding sphere along its full step displacement against a 313-box axis-aligned decomposition of the ISS geometry, so a fast chaser cannot tunnel through thin structure between timesteps. Docking succeeds when the chaser holds within \SI{0.1}{\meter} of the port pose at under \SI{0.5}{\meter\per\second}, within \SI{5}{\degree} of the port attitude, and under \SI{0.5}{\degree\per\second} of body rate. The station model is a NASA-derived ISS asset in its February--May 2015 configuration, on which eight visiting-vehicle ports are reachable targets, spanning all three docking mechanism families: the IDSS port at Harmony forward (PMA-2), the CBM berths at Harmony zenith, Harmony nadir, and Unity nadir, and the SSVP probe-and-drogue ports at Zvezda aft, Poisk zenith, Pirs nadir, and Rassvet nadir. The reward is a weighted sum of position, velocity, attitude, and body-rate errors, each passed through a smoothed pseudo-Huber cost, with large terminal payments for collision, successful docking, and escape; \Cref{sec:experiments-baselines} gives the functional form in \Cref{eq:rl-reward} together with the weights used for data generation and for reinforcement learning training. The attitude and rate terms are gated by distance so that pointing barely matters at range and matters fully at the port, and the position term targets the assigned dock pose rather than the station origin. The world model never trains on the reward.

Observations follow the formulation of \Cref{sec:methods-formulation}. The kinematic channel reports the 13-dimensional relative view through a configurable Gaussian sensor model described in \Cref{sec:experiments}, optionally augmented with a 12-dimensional dock-goal error block. The visual channel is a $512 \times 512$ first-person view from a camera mounted on the chaser's nose looking along the body approach axis with an \SI{82}{\degree} vertical field of view, rendered by a physically based renderer that includes the full-globe textured Earth, starfield, Moon, and epoch-dependent Sun lighting, so the model sees the same lighting variation an approach camera would. Frames are rendered from the recorded simulation states when datasets are written, at the \SI{20}{\hertz} simulation rate.

The two remaining environments trade fidelity for speed. The \texttt{iss-hcw} environment replaces the numerical propagation with Hill--Clohessy--Wiltshire relative dynamics~\cite{Clohessy1960terminal}, adding the gravity-gradient torque, and the \texttt{iss} environment drops orbital mechanics entirely, propagating a rigid-body free-flyer whose translation reduces to a double integrator under thrust.

\section{Experiments}
\label{sec:experiments}

We evaluate the approach on the ISS docking scenario across three sensor noise regimes, comparing world model planning against reinforcement learning baselines and measuring prediction quality, docking performance, generalization to unseen ports, and anomaly detection. This section describes the data generation, training configuration, baselines, and evaluation criteria.

\subsection{Data Generation}
\label{sec:experiments-data}

\begin{figure}[t]
    \centering
    \includegraphics[width=\textwidth]{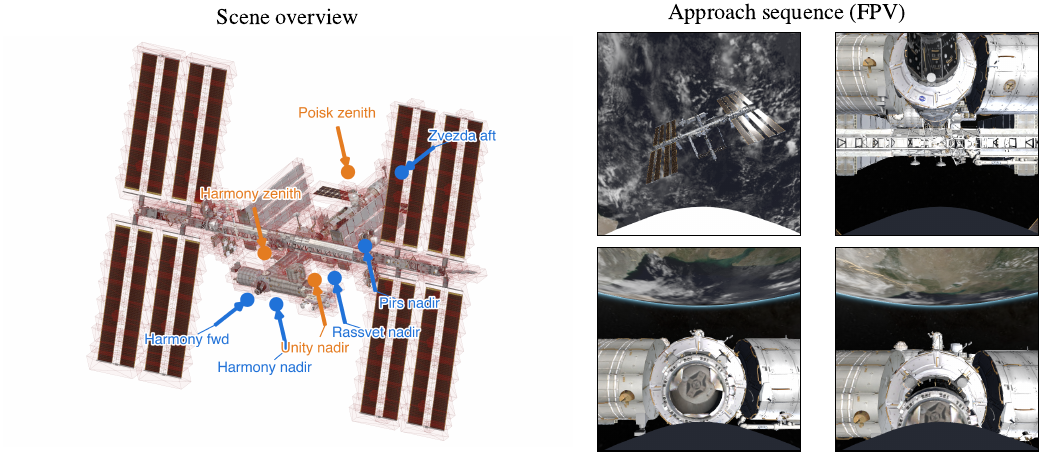}
    \caption{The ISS docking environment. Left: isometric view of the station model overlaid with the 313 axis-aligned collision boxes (red) used for keep-out checking and the goal poses of the eight docking ports. Blue markers label the five ports flown in the training data and orange markers the three ports held out for evaluation; arrows indicate the approach corridors. Right: the chaser's first-person camera view---the visual observation the world model consumes---at four points along a successful docking approach to the Harmony forward port.}
    \label{fig:iss-environment}
\end{figure}

\begin{figure}[t]
    \centering
    \includegraphics[width=\textwidth]{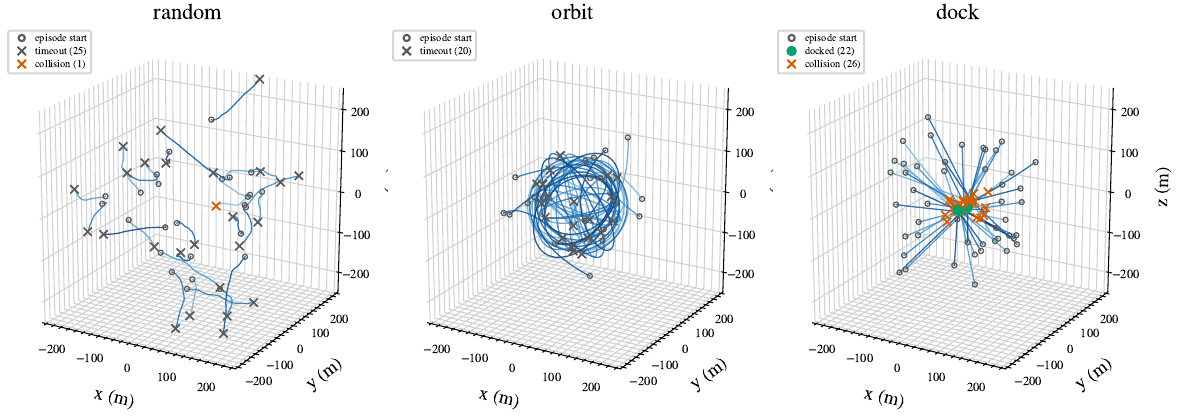}
    \caption{Training trajectories from the published dataset in the station-centered frame, one panel per behavior policy, with episode starts and outcome-coded episode ends. Random walks drift across the domain without goal direction, orbit trajectories circumnavigate the station across a range of radii and plane orientations, and dock trajectories converge on the sampled port poses, with approaches that clip station structure recorded as collisions.}
    \label{fig:dataset-trajectories}
\end{figure}

\Cref{fig:iss-environment} presents the simulation environment. Each episode initializes the chaser at a uniformly sampled radius between \SIrange{100}{225}{\meter} from the station origin with the nose pointed at the station and an epoch drawn uniformly from a seven-day window---which sweeps the solar beta angle and with it the lighting the camera sees---and terminates on docking, collision, departure beyond \SI{750}{\meter}, or the \SI{360}{\second} horizon. Trajectories are generated by three behavior policies, chosen so that the training distribution covers both goal-directed and undirected motion. The \textit{random} policy samples uniform force and torque commands, producing undirected drift that covers the state space far from the station. The \textit{orbit} policy commands a proportional-derivative-tracked circumnavigation of the station at a sampled radius between \SIrange{80}{130}{\meter} of the station, exposing the model to sustained lateral motion and the full range of viewing geometries. The \textit{dock} policy flies a critically damped proportional-derivative approach to a sampled docking port. About half of these approaches meet the contact conditions and the remainder end in recorded collisions with station structure, so the corpus also covers the contact-adjacent failure modes the model must learn to predict. \Cref{fig:dataset-trajectories} shows the training episodes from each policy.

The training split mixes the three policies in a $0.30/0.35/0.35$ ratio and its dock lane flies to five of the eight ports---Harmony forward, Harmony nadir, Zvezda aft, Pirs nadir, and Rassvet nadir. The three remaining ports, Harmony zenith, Poisk zenith, and Unity nadir, appear only in evaluation, giving a held-out generalization test whose goal poses, approach corridors, and visual context the model has never seen. \Cref{tab:datasets} summarizes the generated datasets. Each frame records the true state, the noisy observation vector, the action, the reward, and the rendered first-person camera view, packaged in LeRobot format with normalization statistics computed on the training split alone.

\begin{table}[t]
    \centering
    \caption{Generated datasets. Transition counts are minimum targets; Training generation runs full episodes to until the target transition count is met, so actual transition count for realized dataset may be slightly larger.}
    \label{tab:datasets}
    \begin{tabular}{@{}llll@{}}
        \toprule
        Split & Transitions & Policy mix & Dock ports \\
        \midrule
        train & $500,000$ & random/orbit/dock (0.30/0.35/0.35) & 5 \\
        val & $50,000$ & dock & 8 \\
        \bottomrule
    \end{tabular}
\end{table}

Sensor noise follows one of three models, summarized in \Cref{tab:noise}, for training and evaluation. The behavior policies act on the noisy measurement rather than on the true state, so the recorded action-outcome pairs show the aleatoric transition uncertainty of acting on an observed state rather than the true state, which would result in deterministic dynamics. The \textit{cooperative} model represents differential-GNSS-class relative navigation with a fixed position error budget, as flown between vehicles that exchange carrier-phase measurements. The \textit{non-cooperative} model is based on vision-based navigation, in which position error grows with range and the velocity estimate is coarser. Attitude and rate noise, which are assumed to come from the chaser's own star tracker and gyroscopes, are identical across the two presets. \Cref{tab:noise} lists the values. All noise magnitudes are total root-mean-square errors---the Euclidean norm of the error vector---matching the convention of navigation requirements documents. Finally there is a \textit{no noise} configuration in which no kinematic noise is added, enabling isolation and quantification of world-model reconstruction errors.

\begin{table}[t]
    \centering
    \caption{Sensor noise models. Values are total-RMS errors.}
    \label{tab:noise}
    \begin{tabular}{@{}lcccc@{}}
        \toprule
        Preset & Position & Velocity & Attitude & Body rate \\
        \midrule
        No noise & 0 & 0 & 0 & 0 \\
        Cooperative & \SI{0.05}{\meter} & \SI{0.002}{\meter\per\second} & \SI{5e-5}{\radian} & \SI{1e-5}{\radian\per\second} \\
        Non-cooperative & 1\% of range & \SI{0.03}{\meter\per\second} & \SI{5e-5}{\radian} & \SI{1e-5}{\radian\per\second} \\
        \bottomrule
    \end{tabular}
\end{table}

\subsection{Training Configuration}
\label{sec:experiments-training}

World models train on windows of $H = 8$ context steps and $F = 64$ rollout steps with the procedure of \Cref{sec:methods-training}: AdamW at a learning rate of \num{e-3}, 300-step linear warmup, mixed \texttt{bfloat16} precision, gradient clipping at 1.0, and teacher forcing annealed to zero over the first four epochs. Batch sizes are tuned automatically to fill available GPU memory. {Each model trains for 16 epochs on one NVIDIA H100 GPU with 96\,GB of memory, completing in approximately 24 hours; data generation for the 500{,}000-transition corpus, including rendering, completes in approximately 6 hours on the same hardware.}

\subsection{Baseline Algorithms}
\label{sec:experiments-baselines}

We compare against a model-free reinforcement learning baseline trained directly on the environment: proximal policy optimization (PPO)~\cite{Schulman2017ppo}, in its Stable-Baselines3 implementation~\cite{Raffin2021sb3}. The policy is a multilayer perceptron consuming the normalized 27-dimensional vector observation (the epoch, the noisy relative state view, and the dock-goal error block) and trained separately for each the cooperative and non-cooperative sensor noise models, on the five training-data ports and a start shell of \SIrange{100}{125}{\meter}. The goal-error block is itself an augmentation of the default model-free formulation: without an input identifying the commanded port, a trained policy has no mechanism for generalizing to a new docking task at all, so we condition the baseline on the goal state to make port transfer expressible in the first place.

Obtaining a reinforcement learning baseline that makes progress on this task at all required significant adaptation and tuning of the training configuration. We document the adaptations for reproducibility and because they measure the challenge with implementing the approach. The environment's native configuration defeats temporal-difference learning: at the \SI{20}{\hertz} simulation step the terminal docking bonus sits 7200 steps out, where at $\gamma = 0.995$ it is discounted by $\gamma^{7200} \approx 2 \times 10^{-16}$, a contribution that vanishes against the accumulated per-step shaping in single-precision arithmetic, so the policy see no contribution from successful docks near the end of the horizon. The baseline therefore trains at a \SI{1}{\hertz} control step, integrating the same dynamics over the same \SI{360}{\second} horizon in 360 decisions rather than 7200, the cadence used in prior reinforcement learning docking work~\cite{Broida2019rendezvous, Dunlap2023runtime}, with the keep-out zone made soft, so that collisions are charged per step rather than terminating the episode. Finally, since past works only considered much simpler keep-out zones and restricted dynamics to only orbital (no attitude) dynamics, we had to spend significant effort shaping a reward function that successfully encouraged docking in a correct orientation over loitering at a safe distance.

The reward used for baseline training is, per step,
\begin{equation}
\begin{aligned}
    r_t ={} & w_{\mathrm{pos}}\, h(d_t; \delta_{\mathrm{pos}}, \sigma_{\mathrm{pos}})
    + w_{\mathrm{vel}}\, h(\lVert\mathbf{v}_t\rVert; \delta_{\mathrm{vel}}, \sigma_{\mathrm{vel}})
    + g(d_t) \left[ w_{\mathrm{att}}\, h(\theta_t; \delta_{\mathrm{att}}, \sigma_{\mathrm{att}})
    + w_{\mathrm{rate}}\, h(\lVert\bm{\omega}_t\rVert; \delta_{\mathrm{rate}}, \sigma_{\mathrm{rate}}) \right] \\
    & + w_{\mathrm{prog}}\, \frac{d_t - d_{t-1}}{\sigma_{\mathrm{pos}}}
    + e^{-d_t / \sigma_{\mathrm{prox}}} \left( w_{\mathrm{prox}} + w_{\mathrm{align}}\, e^{-\theta_t / \sigma_{\mathrm{align}}} \right)
    + w_{\mathrm{eff}} \left( \frac{\lVert\mathbf{F}_t\rVert}{F_{\max}} + \frac{\lVert\bm{\tau}_t\rVert}{\tau_{\max}} \right) \\
    & + w_{\mathrm{col}}\, \mathbf{1}_{\mathrm{col}}
    + w_{\mathrm{dock}}\, \mathbf{1}_{\mathrm{dock}}
    + w_{\mathrm{esc}}\, \mathbf{1}_{\mathrm{esc}}
\end{aligned}
\label{eq:rl-reward}
\end{equation}
where $d_t$ is the distance to the assigned port position, $\theta_t$ the attitude error to the port attitude, $\mathbf{v}_t$ and $\bm{\omega}_t$ the relative velocity and body rate, and $\mathbf{F}_t$ and $\bm{\tau}_t$ the commanded force and torque against their actuator limits ($F_{\max} = \SI{1600}{\newton}$, $\tau_{\max} = \SI{2000}{\newton\meter}$). Each error norm passes through the normalized pseudo-Huber loss
\begin{equation}
    h(e; \delta, \sigma) = \frac{\sqrt{e^2 + \delta^2} - \delta}{\sigma},
    \label{eq:pseudo-huber}
\end{equation}
which is quadratic below the knee $\delta$ and has unit far-field slope over the scale $\sigma$, so the far field keeps a constant pull toward the port while the near field stays smooth. The rotational terms are gated by
\begin{equation}
    g(d) = g_\infty + \frac{1 - g_\infty}{1 + (d / d_g)^2}
    \label{eq:rotation-gate}
\end{equation}
so that pointing matters fully at the port and fractionally at range. The potential-based progress term~\cite{Dunlap2023runtime} pays for range closed when it is closed---without it, tuned agents trained for 20 million steps learned to stop escaping but never to approach. The proximity and alignment bonuses are a smooth, discount-reachable precursor of the docking event, paying up to $w_{\mathrm{prox}} + w_{\mathrm{align}}$ per step for sitting on the port pointed correctly. The indicators fire on collision, docking, and departure beyond the domain boundary. \Cref{tab:rl-reward} lists the parameter values.

\begin{table}[t]
    \centering
    \small
    \begin{minipage}[t]{0.63\textwidth}
    \centering
        \caption{Reward parameters used for reinforcement learning baseline training (\Cref{eq:rl-reward}).}
        \label{tab:rl-reward}
        \begin{tabular}{@{}lll@{}}
            \toprule
            Term & Weight & Shaping \\
            \midrule
            Position & $w_{\mathrm{pos}} = -0.9$ & $\delta_{\mathrm{pos}} = \SI{1}{\meter}$, $\sigma_{\mathrm{pos}} = \SI{225}{\meter}$ \\
            Velocity & $w_{\mathrm{vel}} = -0.05$ & $\delta_{\mathrm{vel}} = \SI{0.1}{\meter\per\second}$, $\sigma_{\mathrm{vel}} = \SI{5}{\meter\per\second}$ \\
            Attitude & $w_{\mathrm{att}} = -0.2$ & $\delta_{\mathrm{att}} = \SI{0.05}{\radian}$, $\sigma_{\mathrm{att}} = \pi\,\si{\radian}$ \\
            Body rate & $w_{\mathrm{rate}} = -0.02$ & $\delta_{\mathrm{rate}} = \SI{0.005}{\radian\per\second}$, $\sigma_{\mathrm{rate}} = \SI{0.05}{\radian\per\second}$ \\
            Rotation gate & --- & $g_\infty = 0.5$, $d_g = \SI{25}{\meter}$ \\
            Progress & $w_{\mathrm{prog}} = -2.0$ & shares $\sigma_{\mathrm{pos}}$ \\
            Proximity & $w_{\mathrm{prox}} = +0.5$ & $\sigma_{\mathrm{prox}} = \SI{1}{\meter}$ \\
            Alignment & $w_{\mathrm{align}} = +0.5$ & $\sigma_{\mathrm{align}} = \SI{0.1}{\radian}$ \\
            Effort & $w_{\mathrm{eff}} = -0.05$ & normalized by actuator limits \\
            Collision & $w_{\mathrm{col}} = -2$ per step & non-terminating \\
            Dock success & $w_{\mathrm{dock}} = +1000$ & terminating \\
            Escape & $w_{\mathrm{esc}} = -1000$ & terminating \\
            \bottomrule
        \end{tabular}
    \end{minipage}\hfill
    \begin{minipage}[t]{0.33\textwidth}
    \centering
        \caption{PPO baseline configuration.}
        \label{tab:rl-final}
        \begin{tabular}{@{}ll@{}}
            \toprule
            Parameter & Value \\
            \midrule
            Total environment steps & 25M \\
            Parallel environments & 32 \\
            Start shell & \SIrange{100}{125}{\meter} \\
            Learning rate & \num{2.2e-4} \\
            Discount $\gamma$ & 0.997 \\
            Network & $256 \times 3$ \\
            Batch size & 1024 \\
            Rollout length & 1024 \\
            GAE $\lambda$ & 0.945 \\
            Clip range & 0.1 \\
            Epochs per update & 20 \\
            Value loss coefficient & 0.67 \\
            Entropy coefficient & \num{e-3} \\
            Target KL divergence & 0.02 \\
            \bottomrule
        \end{tabular}
    \end{minipage}
\end{table}

Hyperparameters were selected by Bayesian optimization with Hyperband early termination~\cite{Li2018hyperband}; \Cref{tab:rl-final} lists the frozen configuration, with the entropy coefficient floored at \num{e-3} and a target KL divergence of 0.02 so that exploration does not collapse over a long run. The policy trains for at most 25 million environment steps under each noise preset. The cooperative policy is evaluated at its best-validation-return checkpoint, at 22.5 million steps.

The world model plans with model predictive path integral (MPPI) control~\cite{Williams2017mppi}. At each replanning step the planner perturbs its nominal action sequence with Gaussian noise to produce $K$ candidate sequences, rolls each through the world model in latent space, scores the decoded state predictions against the docking objective, and updates the nominal sequence with the softmax-weighted average
\begin{equation}
    \mathbf{a}^\star = \sum_{k=1}^{K} w_k\, \mathbf{a}_k \qquad
    w_k = \frac{\exp\!\left(R_k / \lambda\right)}{\sum_{j} \exp\!\left(R_j / \lambda\right)}
    \label{eq:mppi}
\end{equation}
where $R_k$ is the predicted return of candidate $k$ and $\lambda$ is a temperature. Because the world model rolls all $K$ candidates as one batch with a shared encoded context and a key--value cache, a full replanning step is a single batched forward pass. The planner executes a short chunk of the optimized sequence open-loop, then re-encodes the new observation and replans. We use a planning horizon of 64 steps, $K = 128$ candidate sequences, a temperature of $\lambda = 0.3$, and execute 8-step chunks between replans.

\subsection{Evaluation Criteria}
\label{sec:experiments-evaluation}

We evaluate along four axes:

\begin{enumerate}
    \item \textbf{Prediction quality.} Open-loop state reconstruction error and image reconstruction quality (mean-squared error and peak signal-to-noise ratio) on held-out trajectories, as a function of training epoch and of prediction horizon, under each of the three noise presets. Comparing the no-noise, cooperative, and non-cooperative presets isolates how much sensor noise---rather than model capacity---limits predictive accuracy, and how noise degrades position and velocity prediction differently.
    \item \textbf{Docking with in-distribution goals.} Each method flies 50 rollouts per port for the five ports in the training data under the cooperative noise preset. Because the full contact conditions of \Cref{sec:methods-env} proved out of reach for initial reinforcement learning baseline attempts, we report success at graduated requirement levels: an attempt passes the position-only definition at tolerance $d$ if its closest approach to the goal pose comes within $d$, and passes the full definition if the velocity, attitude, and body-rate bounds also hold at that approach. Any episode in which the chaser contacts station structure is counted as a failure under every definition. We report the position-only rate at $d = \SI{0.5}{\meter}$ per port, sweep $d$ from the \SI{0.1}{\meter} contact gate outward to trace each method's precision profile under both definitions, and report the collision rate per port alongside.
    \item \textbf{Docking with out-of-distribution goals.} The same protocol on Harmony zenith, Poisk zenith, and Unity nadir, which no policy visited during training. Reinforcement learning policies must generalize zero-shot through their goal-error input; the world model receives only a new goal pose in the planner's objective.
    \item \textbf{Anomaly detection.} We render evaluation episodes in which an uninvolved visiting vehicle, a second Dragon spacecraft, is present in the environment, and measure whether the world model's predictive uncertainty flags the anomaly. We run 20 rollouts including the anomalous docked vehicle (even if not physically possible) and report the detection rate defined by the uncertainty exceeding a threshold calibrated on 25 nominal docking episodes.
\end{enumerate}

\section{Results}
\label{sec:results}

\subsection{Training and Prediction Quality}
\label{sec:results-training}

\begin{figure}[t]
    \centering
    \begin{subfigure}[t]{0.49\textwidth}
        \centering
        \includegraphics[width=\textwidth]{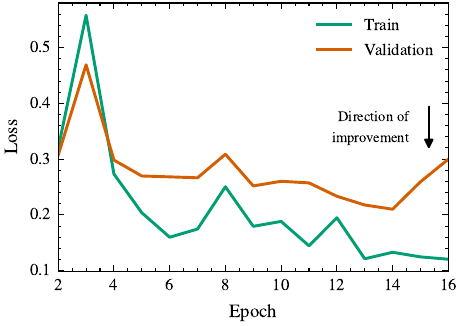}
        \caption{World model training and validation loss. We note that we stop training as validation loss starts to climb, and use the lowest validation model loss checkpoint for evaluations.}
        \label{fig:training-loss-wm}
    \end{subfigure}\hfill
    \begin{subfigure}[t]{0.49\textwidth}
        \centering
        \includegraphics[width=\textwidth]{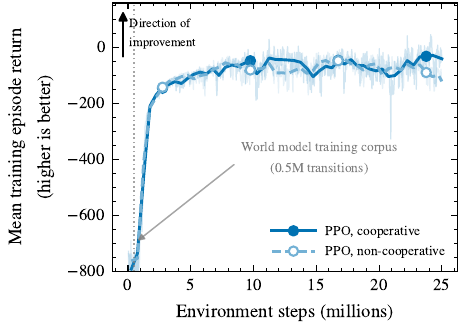}
        \caption{Reinforcement learning baseline training return.}
        \label{fig:training-loss-rl}
    \end{subfigure}
    \caption{Training curves. (a) Total training loss by epoch for the flow-matching OWM architecture and the Dreamer-style posterior-correction baseline. (b) Mean training episode return over environment steps for the PPO baseline under cooperative and non-cooperative sensor noise, averaged in bins of $5\times10^5$ steps over the raw rolling mean; the non-cooperative run is drawn dashed . The dotted line marks the equivalent size of the world model's training corpus on the same axis.}
    \label{fig:training-loss}
\end{figure}

\begin{figure}[t]
    \centering
    \includegraphics[width=\textwidth]{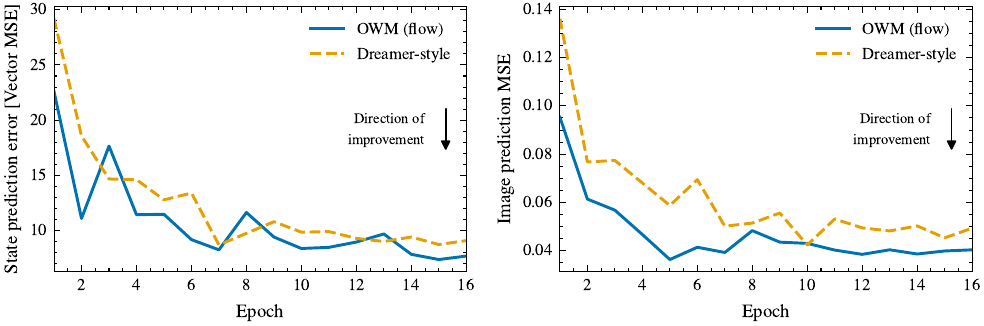}
    \caption{World model validation performance by training epoch. Left: state reconstruction error on held-out trajectories under fully autoregressive rollout. Right: image reconstruction loss on the same rollouts.}
    \label{fig:wm-learning-curves}
\end{figure}

\begin{figure}[t]
    \centering
    \includegraphics[width=\textwidth]{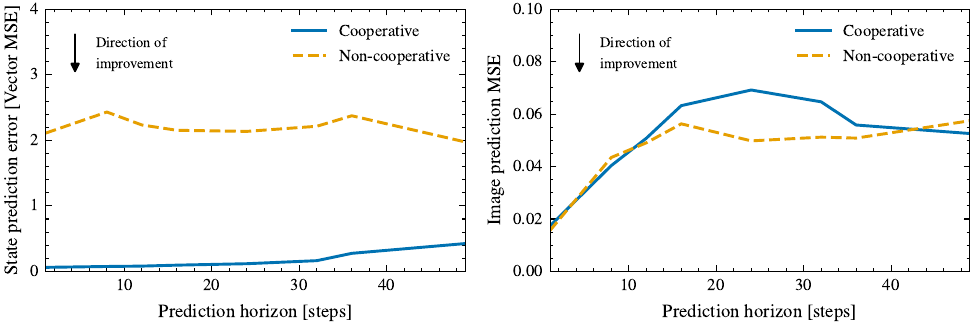}
    \caption{Prediction error against rollout horizon for each sensor noise preset. Left: state prediction error. Right: image prediction loss. Non-cooperative range-proportional noise degrades long-horizon state prediction more than image predictions.}
    \label{fig:horizon-error}
\end{figure}

\Cref{fig:training-loss} presents the training curves for the world model variants and reinforcement learning baselines, and \Cref{fig:wm-learning-curves} presents validation reconstruction quality by epoch. The reinforcement learning baseline (\Cref{fig:training-loss-rl}) makes most learning progress within the first 5 million steps and improves slowly thereafter; the cooperative policy is evaluated at its best-performing checkpoint on the held-out validation episodes (22.5 million steps). {Both world model variants converge within 16 epochs. We find that the flow-matching architecture reaches lower validation prediction errors---for both state and image prediction---compared to the Dreamer-style baseline, while using fewer trainable parameters.}

\Cref{fig:horizon-error} shows prediction error as a function of rollout horizon under each noise preset. We find that the state prediction error is significantly degraded in the non-cooperative setting; it is consistently poorer across the prediction horizon. We expect that this is due to the noise perturbations being applied directly to the state observations. In general, these perturbations greatly increase the MSE of our kinematic state predictions, but image prediction quality is nearly insensitive to the kinematic noise preset, which is expected: the camera channel is noise-free in all presets.


\subsection{Docking Performance}
\label{sec:results-docking}

\begin{figure}[t]
    \centering
    \includegraphics[width=\textwidth]{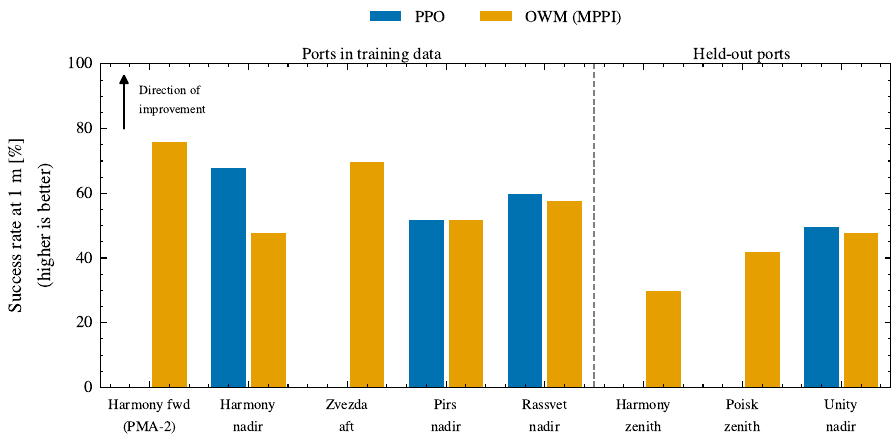}
    \caption{Docking success rate by port and algorithm under cooperative sensor noise, over 50 rollouts per port. An attempt succeeds if its closest approach to the port comes within \SI{1}{\meter} and the chaser never contacts station structure during the episode. Ports left of the divider appear in the training data; Harmony zenith, Poisk zenith, and Unity nadir are held out.}
    \label{fig:docking-coop}
\end{figure}

\begin{figure}[t]
    \centering
    \includegraphics[width=\textwidth]{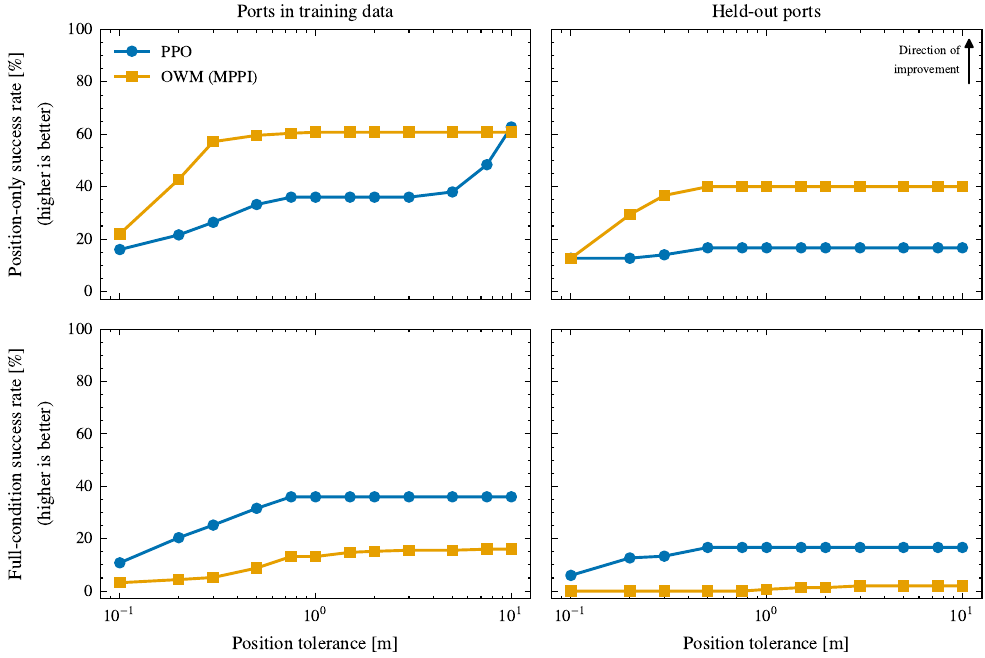}
    \caption{Collision-voided success rate against position tolerance under cooperative sensor noise, pooled over the ports in the training data (left) and the held-out ports (right), over 50 rollouts per port. Top: position-only success, in which the closest approach comes within the tolerance. Bottom: the full contact conditions, which additionally bound the velocity, attitude error, and body rate at that approach. Curves are shown for the world-model planner and the PPO baseline. The two rows expose the central trade-off; the planner leads on position-only success through reliable acquisition, while PPO leads on the full contact conditions through its slower, better-pointed terminal approach.}
    \label{fig:success-vs-tolerance}
\end{figure}

\begin{figure}[t]
    \centering
    \includegraphics[width=\textwidth]{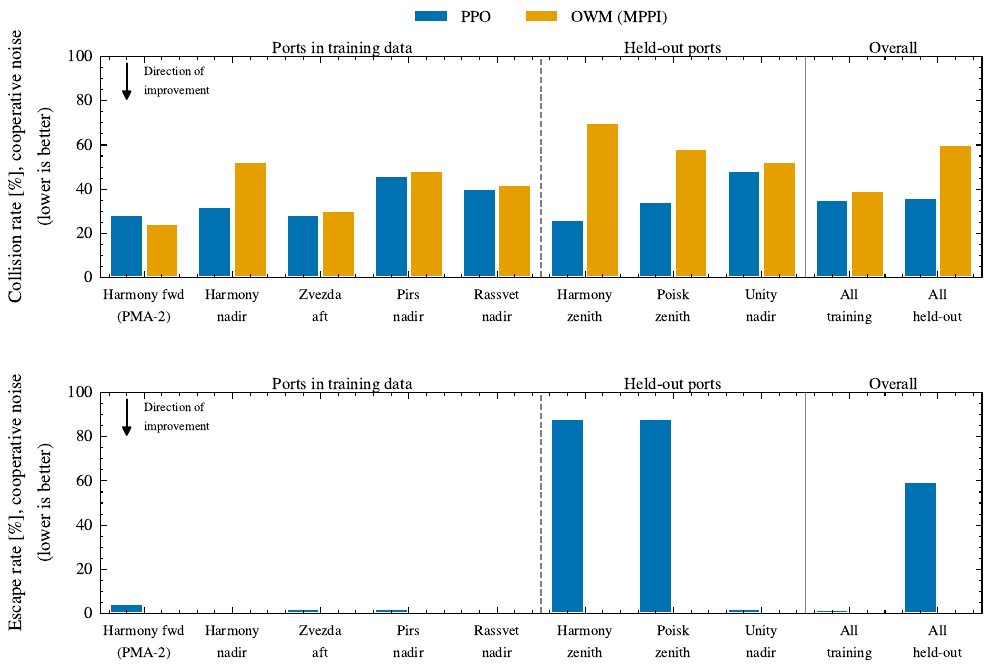}
    \caption{Collision rate (top) and escape rate (bottom) by port and algorithm under cooperative sensor noise, over 50 rollouts per port. The collision rate is the fraction of rollouts in which the chaser entered a station keep-out box at any point; the escape rate is the fraction in which the chaser left the \SI{750}{\meter} operational domain. The world-model planner never escapes, so every one of its rollouts presses a full approach and attempts a dock.}
    \label{fig:collision-rate}
\end{figure}


\Cref{fig:docking-coop} presents docking success rates by port under cooperative noise, with the five ports in the training data and the three held-out ports (Harmony zenith, Poisk zenith, and Unity nadir), which no training trajectory visited. Two patterns stand out. First, given only the port's goal pose in its planning objective and minimal controller tuning, the world-model planner is comparable to PPO for every port and docks at 4 ports the PPO policy never does. The world-model planner's collision-voided success at the \SI{1}{\meter} tolerance on the training ports is $61\%$ against $36\%$ for PPO. It docks at Harmony forward ($76\%$) and Zvezda aft ($70\%$), two training ports the RL policy never reaches (its approaches stall \SIrange{10}{12}{\meter} out), while PPO docks in earnest only at the three nadir-facing training ports ($68\%$, $60\%$, and $52\%$ at Harmony, Rassvet, and Pirs nadir). Second, and more consequentially, the held-out ports expose a sharp difference in out-of-distribution generalization. The planner, which receives only a new goal pose and is otherwise unchanged, generalizes to all three: $48\%$ at Unity nadir, $42\%$ at Poisk zenith, and $30\%$ at Harmony zenith. PPO The baseline, despite its goal-state conditioning, generalizes to exactly one held-out port, Unity nadir ($50\%$), which shares the nadir approach direction of the Pirs and Rassvet ports in the training set, and fails outright at the two zenith ports, whose approach geometry resembles no training task. Reinforcement learning, that is, transfers only where a held-out port looks like a task it has already trained on; the world model transfers to every port it is pointed at, despite having seen none of them.

\Cref{fig:success-vs-tolerance} sweeps the position tolerance of the success definition at two grades of strictness. Under position-only success (top row), the planner outperforms the baseline handily at every tolerance, on training and held-out ports alike, reflecting its reliable acquisition. Under the full contact conditions (bottom row), which additionally bound the velocity, attitude error, and body rate at closest approach, the ordering reverses: PPO's slower, better-pointed terminal approaches meet all four conditions in $36\%$ of training-port rollouts at \SI{1}{\meter} against the planner's $13\%$ (concentrated at Harmony forward, $66\%$), and $11\%$ against $3\%$ at the strict \SI{0.1}{\meter} contact gate. We attribute much of this gap to tuning effort rather than to the paradigm: the RL reward shaping was iterated over many training runs to produce slow, pointed arrivals, whereas the MPPI cost received far less tuning and favors a collision-free approach over arriving pointed. More careful weighting of the terminal attitude and body-rate costs should improve attitude-rate convergence and lift the planner's success under the full contact conditions.

\Cref{fig:collision-rate} reports the collision rate (top panel) and escape rate (bottom panel) at each port; the two panels must be read together. In isolation the planner's collision rate is the higher one ($39\%$ of training-port rollouts against PPO's $35\%$, and $60\%$ against $36\%$ on the held-out ports), but the escape rates show why. The planner never leaves the operational domain: $0\%$ of rollouts across all eight ports, versus the baseline's $88\%$ at each zenith port. Rollouts that escape before any close approach can never register a collision, so the baseline's low held-out collision rate reflects approaches it never made. The planner instead presses a full approach on every rollout, attempting many more docks precisely where the station geometry is tightest, and its residual failure mode is contact with structure on the zenith approaches rather than lost acquisition. Reducing this collision rate, whether through the cost tuning above or by feeding the model's predictive uncertainty back into the planner as a runtime safety signal, remains open work.

\subsection{Anomaly Detection}
\label{sec:results-anomaly}


\begin{figure}[t]
  \centering
  \begin{subfigure}[t]{0.48\linewidth}
    \centering
    \includegraphics[width=\linewidth]{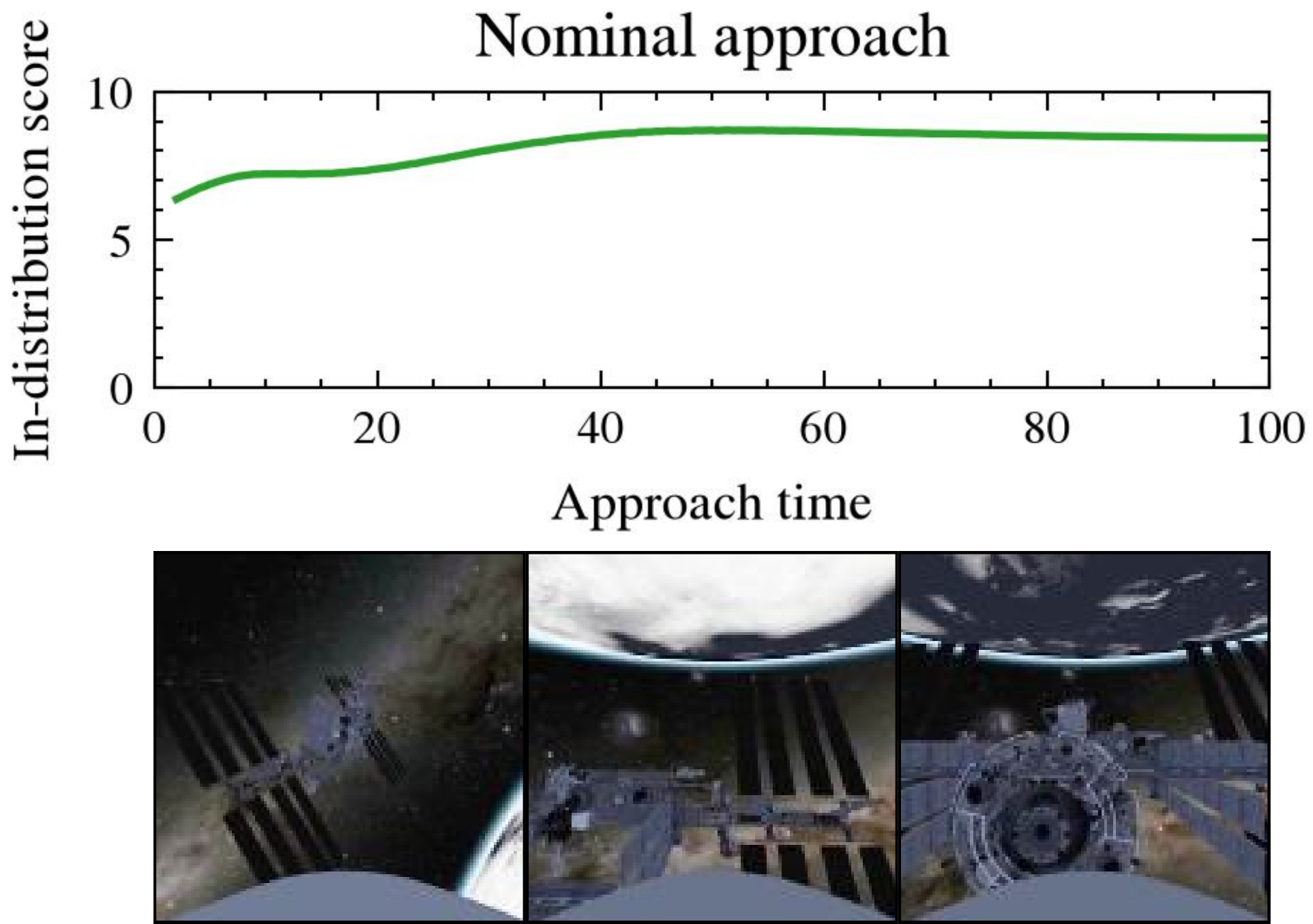}
    \caption{A nominal approach: the in-distribution score remains high throughout, indicating the world is evolving as the model predicted.}
    \label{fig:example-short-ind}
  \end{subfigure}
  \hfill
  \begin{subfigure}[t]{0.48\linewidth}
    \centering
    \includegraphics[width=\linewidth]{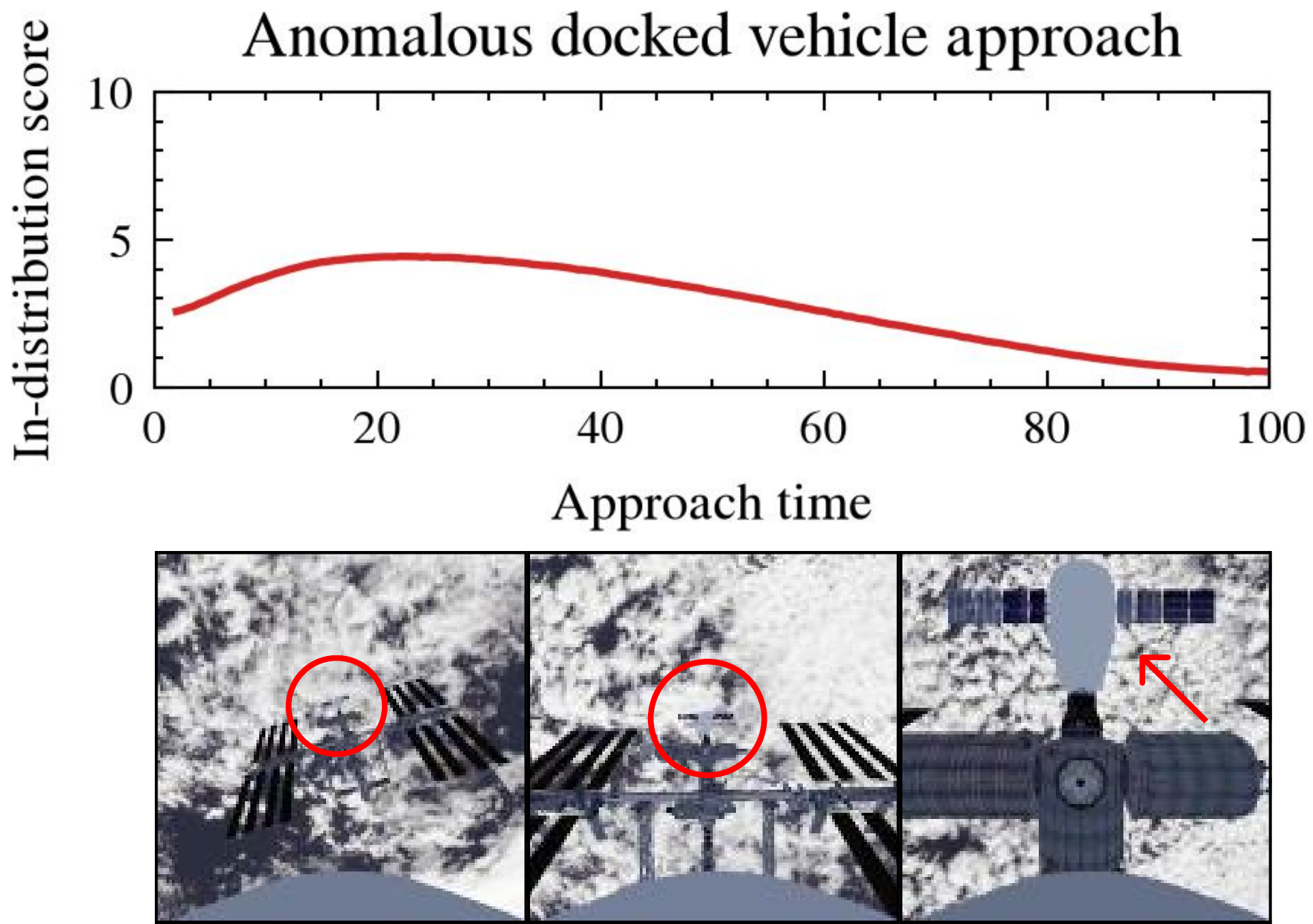}
    \caption{An approach with an anomalous docked vehicle: the model, which never saw a docked vehicle in training, cannot predict its presence, and the score drops as the vehicle grows in the perceptual field.}
    \label{fig:example-short-ood}
  \end{subfigure}
\caption{In-distribution scores as an anomaly signal on two representative approaches.}
\label{fig:anomaly}
\end{figure}

\begin{figure}[t]
  \centering
  \includegraphics[width=0.35\linewidth]{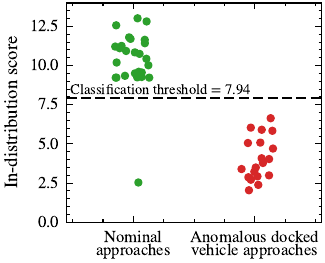}
  \caption{Maximum in-distribution score per approach over the anomaly detection dataset. Approaches with an anomalous docked vehicle score consistently lower, and a scalar threshold classifies $98\%$ of approaches correctly.}
  \label{fig:aggregate-classifier}
\end{figure}

Finally, we evaluate the world model's predictive uncertainty as a runtime anomaly monitor. \Cref{fig:anomaly} shows two representative episodes, and \Cref{fig:aggregate-classifier} the aggregated results across the full anomaly detection dataset. The world model, having never seen a second visiting vehicle during training, predicts a nominal scene, and the disagreement between prediction and observation concentrates precisely on the anomalous object. The model correctly classifies $20$ approach sequences including an anomalous docked Dragon spacecraft from $25$ approach sequences without any anomalous spacecraft with a classification accuracy of $98\%$. 

\section{Conclusions}
\label{sec:conclusions}

This paper introduces a world model approach to spacecraft rendezvous and proximity operations. We present Out-of-this-World-Model, a transformer-based architecture that fuses camera imagery and kinematic state measurements into a shared latent space and predicts distributions over future observations with a one-step flow-matching head, and we present AstroJAX, a JAX-based astrodynamics framework whose massively parallel GPU simulation generated the 500{,}000-transition training corpus used in this work. Applied to the problem of a capsule docking with the ISS, the learned world model composes with model predictive path integral control to reach berthing ports both included and excluded from the training distribution, where it more than doubles the docking success rate of reinforcement-learning baselines on held-out ports (40\% versus 17\%), succeeds at ports the baselines fail to acquire at all, and never leaves the operational domain. The same model's predictive uncertainty serves as a runtime anomaly monitor, correctly classifying approach sequences with and without previously unseen vehicles 98\% of the time. Together, these results represent a step toward autonomous spacecraft that reason about the consequences of their actions before executing them.

Promising avenues for future work include more careful tuning of the MPPI cost weights, whose terminal attitude and body-rate terms received far less attention than the reinforcement-learning reward shaping and which we expect to both reduce the planner's collision rate and close its remaining gap under the full contact conditions; leveraging the anomaly detection capability more directly for collision mitigation, so that rising predictive uncertainty tempers approach aggressiveness before contact occurs; applying the learned system model paradigm to pattern-of-life characterization for space domain awareness, where deviations from a satellite's predicted behavior signal a change in operational mode; and integration of world-model based planners with safety-aware planning techniques to further improve docking success rate.

\section*{ACKNOWLEDGMENTS}

{Duncan Eddy is supported by Stanford Institute for Human-Centered AI. Grace Kim is supported by the Fannie and John Hertz Foundation and the National Defense Science and Engineering Graduate (NDSEG) Fellowship Program. Specifically, this material is based upon work supported by the Air Force Office of Scientific Research under award number FA9550-25-C-B010 in the amount of currently negotiated tuition and stipend rates.}

\bibliographystyle{unsrtnat}

\begingroup
\renewcommand{\section}[2]{\par\noindent\textbf{\MakeUppercase{References}}\par}%
\bibliography{references}
\endgroup

\end{document}